\documentclass[letterpaper, 10 pt, conference]{ieeeconf}

\IEEEoverridecommandlockouts                              
\usepackage{graphics} 
\usepackage{amsmath} 
\usepackage{amssymb} 
\usepackage{adjustbox}
\usepackage{tabularx}
\usepackage{booktabs}
\usepackage{algorithm}
\usepackage{algpseudocode}
\usepackage{microtype} 
\usepackage{graphicx}
\usepackage{siunitx}
\usepackage{newtxtext,newtxmath}
\usepackage[table]{xcolor}
\usepackage{placeins}
\usepackage{verbatim}
\usepackage{amsmath}
\usepackage{amsfonts}
\usepackage{graphicx}
\usepackage{array}
\usepackage{float}
\usepackage{makecell} 
\usepackage{multirow} 
\usepackage{booktabs}
\usepackage{caption} 
\usepackage{cite}
\usepackage{tocloft}
\usepackage{subcaption}
\usepackage{pifont}
\usepackage[hidelinks]{hyperref}

\newcommand{\cmark}{\ding{51}}
\newcommand{\xmark}{\ding{55}}

\title{\LARGE \bf
Event-Adaptive Motion Planning with Distilled Vision-Language Model in Safety-Critical Situations
}

\author{Zhenwei Huang$^{1}$, Changsheng You$^{1,\ast}$, Shuai Wang$^{2,\ast}$, Chao Zhou$^{1}$, Wei Xu$^{3}$, and Yi Gong$^{1}$%
\thanks{This work was supported in part by the National Natural Science Foundation of China under Grants 62571227 and U25A20389, in part by the Program under Grant 2023QN10X152, in part by the Shenzhen Science and Technology Program under Grant RCYX20231211090206005, in part by the CAS-TUBITAK Joint Call under the International Partnership Program of the Chinese Academy of Sciences under Grant 321GJHZ2025118MI, and in part by the Special Funds for the Cultivation of Guangdong College Students' Scientific and Technological Innovation under Grant pdjh2026c11014.}%
\thanks{$^{\ast}$Corresponding authors: Changsheng You and Shuai Wang.}%
\thanks{$^{1}$Zhenwei Huang, Changsheng You, Chao Zhou, and Yi Gong are with the Southern University of Science and Technology (e-mail: 12532510@mail.sustech.edu.cn; youcs@sustech.edu.cn; zhouchao2024@mail.sustech.edu.cn; gongy@sustech.edu.cn). $^{2}$Shuai Wang is with the Shenzhen Institute of Advanced Technology, Chinese Academy of Sciences (e-mail: s.wang@siat.ac.cn). $^{3}$Wei Xu is with Manifold Tech Limited (e-mail: xuwei@manifoldtech.cn).}%
}

\begin{document}

\maketitle
\thispagestyle{empty}
\pagestyle{empty}

%%%%%%%%%%%%%%%%%%%%%%%%%%%%%%%%%%%%%%%%%%%%%%%%%%%%%%%%%%%%%%%%%%%%%%%%%%%%%%%%
\begin{abstract}

Robot navigation in safety-critical scenarios faces significant challenges from unforeseen semantic events, where collisions arise primarily from the unpredictable behaviors of dynamic agents rather than unseen objects. While large vision-language models (VLMs) offer remarkable capabilities in commonsense reasoning, frequently invoking them within the continuous control loop introduces severe computational latency, fundamentally destabilizing physical execution. To address these challenges, we propose event-adaptive motion planning (EAMP), an efficient framework for VLM-based robot navigation. Specifically, a prompt-configurable semantic event trigger (PC-SET) selectively activates semantic intervention by continuously monitoring short temporal clips for behavioral anomalies. Upon triggering, an event-triggered distilled SemNav-VLM, fine-tuned via physically verified semantic distillation, maps detected anomalies into discrete strategy-level decisions. Subsequently, a semantic model predictive control (SMPC) module translates these strategies into dynamic reconfigurations of optimization objectives and geometric references. Extensive experiments in safety-critical logistics scenarios demonstrate that EAMP effectively aligns high-level reasoning with low-level control, significantly improving dynamic safety margins over existing baselines while preserving real-time efficiency.

\end{abstract}

\nocite{11218149}
\nocite{10931823}
\nocite{wang2025openbench}
\nocite{sha2023languagempc}
\nocite{11128826}
\nocite{wen2024dilu}
\nocite{10036019}
\nocite{yu2021model}
\nocite{huang2024drivegpt}
\nocite{baumann2025enhancing}
\nocite{saha2025system}
\nocite{chen2024ocp}
\nocite{dosovitskiy2017carla}
\nocite{shao2024lmdrive}
\nocite{tian2024drivevlm}
\nocite{tanwani2020rilaas}

%--------------------------------------------------
\begin{table*}[!t]
\vspace{3.5pt}
\centering
\caption{Comparison of self-adaptive semantic navigation and control schemes.}
\label{tab:comparison_styleA}

\small
\setlength{\tabcolsep}{4.5pt}
\renewcommand{\arraystretch}{1.15}

\resizebox{\textwidth}{!}{%
\begin{tabular}{l c c c c c c c}
\toprule

\multirow{2}{*}{\textbf{Method}}
& \multicolumn{2}{c}{\textbf{Model Coordination}}
& \multicolumn{3}{c}{\textbf{Perception}}
& \multicolumn{2}{c}{\textbf{Control}} \\
\cmidrule(lr){2-3} \cmidrule(lr){4-6} \cmidrule(lr){7-8}

& \makecell[c]{\textbf{Hierarchical}\\\textbf{Trigger}}
& \makecell[c]{\textbf{Event-driven}\\\textbf{Activation}}
& \makecell[c]{\textbf{Multi-modal}\\\textbf{Perception}}
& \makecell[c]{\textbf{Behavior-level}\\\textbf{Awareness}}
& \makecell[c]{\textbf{Adaptation}\\\textbf{Strategy}}
& \makecell[c]{\textbf{Control}\\\textbf{Interface}}
& \makecell[c]{\textbf{Kinematic}\\\textbf{Safety}} \\
\midrule

RDA~\cite{10036019}           & \xmark & \xmark & \xmark & \xmark & --          & Direct MPC        & \cmark \\
OnBoard-LLM~\cite{baumann2025enhancing}   & \cmark & \xmark & \xmark & \xmark & --          & Parameter Tuning   & \cmark \\
LanguageMPC~\cite{sha2023languagempc}   & \cmark & \xmark & \xmark & \cmark & --          & Parameter Tuning   & \cmark \\
HR-MPC~\cite{11128826}    & \cmark & \xmark & \cmark & \xmark & --          & Parameter Tuning   & \cmark \\
VLMPC~\cite{zhao2024vlmpc}         & \xmark & \xmark & \cmark & \xmark & --          & Action Sampling    & \cmark \\
AdaDrive~\cite{Zhang_2025_ICCV}      & \cmark & \cmark & \cmark & \xmark & Retraining  & Direct Action      & \xmark \\
OCP~\cite{chen2024ocp}           & \cmark & \cmark & \cmark & \xmark & Rule-based  & Waypoint Guidance  & \cmark \\
\textbf{EAMP (Ours)}
             & \textbf{\cmark} & \textbf{\cmark} & \textbf{\cmark} & \textbf{\cmark}
             & Prompt-driven & Parameter Tuning & \textbf{\cmark} \\

\bottomrule
\end{tabular}%
}

\vspace{3pt}
\footnotesize{
\cmark~explicit support \quad \xmark~not supported \quad --~not applicable
}
\vspace{-17pt}
\end{table*}
%--------------------------------------------------

%%%%%%%%%%%%%%%%%%%%%%%%%%%%%%%%%%%%%%%%%%%%%%%%%%%%%%%%%%%%%%%%%%%%%%%%%%%%%%%%
\section{Introduction}

Achieving high efficiency while maintaining safety remains a significant challenge for robot navigation, particularly in safety-critical scenarios such as sudden cut-ins and imminent collisions. The sparse distribution of these corner cases in real-world environments leads to a severe scarcity of relevant training data~\cite{11218149}. Consequently, when encountering these unfamiliar situations, autonomous systems often fail to respond effectively, resulting in overly conservative behaviors such as unnecessary detours, emergency braking, or complete immobilization~\cite{10931823}. These limitations pose a major barrier to the widespread deployment of robots in human-centric applications, including logistics and inspection tasks~\cite{wang2025openbench}.

%--------------------------------------------------
\begin{figure}[t]
  \centering
  \includegraphics[width=\linewidth]{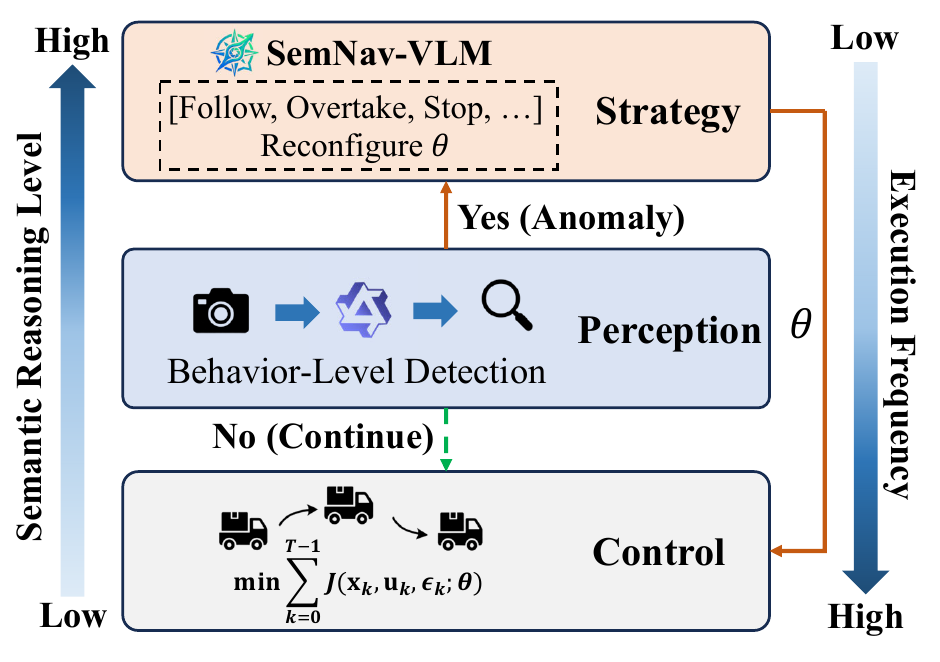}
  \caption{Overview of event-adaptive motion planning.}
  \label{fig:intorpic}
  \vspace{-10pt}
\end{figure}
%--------------------------------------------------

To address this issue, large vision-language models (VLMs) offer a promising solution by leveraging their commonsense reasoning capabilities to recognize corner-case scenarios. However, directly embedding raw VLM outputs into the control loop introduces severe computational latency, which critically destabilizes physical execution. Existing navigation schemes~\cite{sha2023languagempc, 11128826, wen2024dilu} fail to address the above challenges, as they overlook the end-to-end impact of high-level VLM inference on low-level robot control. Specifically, classical optimization-based methods~\cite{10036019, yu2021model} lack semantic understanding, while autoregressive end-to-end models~\cite{huang2024drivegpt} remain computationally prohibitive for high-frequency applications. Although recent decoupled architectures~\cite{baumann2025enhancing, saha2025system} reduce computational overhead, they rely on rigid periodic model invocations. More recently, opportunistic collaborative planning (OCP)~\cite{chen2024ocp} advances this direction by triggering a large vision model (LVM) only upon detecting unknown objects. Nevertheless, this strategy implies that different behaviors of a single entity are treated as equally significant, rendering the system vulnerable to rare but critical behaviors of common agents (e.g., sudden pedestrian jaywalking).

To fill this research gap, we propose in this paper an \texttt{Event Adapter} approach for VLM-based robot navigation in safety-critical situations.
As illustrated in Fig.~\ref{fig:intorpic}, our method initiates behavior-level anomaly detection by prompting a lightweight (e.g., $\leq 3$\,B) onboard VLM. 
Upon an event trigger, it employs a distilled VLM to reconfigure the motion planning parameters, adapting the controller to handle the current corner case. This architecture is referred to as the event-adaptive motion planning (EAMP) framework.
In particular, the specialized VLM, termed as SemNav-VLM, 
is fine-tuned on a \emph{behavior-annotated} vision-parameter dataset 
collected in the Car Learning to Act (CARLA)~\cite{dosovitskiy2017carla}, 
which captures diverse safety-critical interaction events 
and their corresponding control adaptations.
This approach contrasts with existing adaptive navigation schemes that either switch between discrete heuristic planners \cite{10931823, wen2024dilu} or adopt end-to-end general VLMs \cite{shao2024lmdrive, tian2024drivevlm} that frequently generate low-resolution control trajectories.
We conduct experiments in CARLA using the Robot Operating System (ROS). Experimental results demonstrate the superiority of the proposed EAMP over the RDA controller \cite{10036019}, periodical collaboration strategy (PCS) \cite{tanwani2020rilaas}, and OCP \cite{chen2024ocp} in safety-critical navigation scenarios.

Our main contributions are summarized as follows:

\begin{itemize}

\item Propose \textbf{EAMP}, which enables self-adaptive navigation while preserving control stability and interpretability.

\item Develop a prompt-configurable semantic event trigger \textbf{(PC-SET)} that performs behavior-level anomaly detection via a lightweight onboard VLM, activating semantic adaptation only when interaction semantics change.

\item Design a semantic model predictive control \textbf{(SMPC)} module driven by SemNav-VLM to translate strategy-level decisions into structured objective reconfigurations.

\item Build a long-tail smart logistics benchmark in CARLA to validate the robustness and efficiency of EAMP in safety-critical urban scenarios.

\end{itemize}

\section{Related Work} 
\label{sec:related}

Recent works integrate VLMs to enhance autonomous navigation performance and other embodied perception tasks~\cite{kou2026beamvlm}. 
As summarized in Table~\ref{tab:comparison_styleA}, existing paradigms differ in model coordination, perception abstraction, and control integration. 
Specifically, classical optimization-based planning systems and interaction-aware hierarchical planners (e.g., EPSILON~\cite{ding2021epsilon}) ensure kinematic feasibility and efficiency through structured optimization, while they do not leverage language-level semantic reasoning for behavior adaptation.
In contrast, unified perception-to-planning frameworks such as UniAD~\cite{hu2023planning} and vision-language-action models including RT-2~\cite{zitkovich2023rt}, LMDrive~\cite{shao2024lmdrive}, and DiLu~\cite{wen2024dilu} incorporate multimodal reasoning into policy generation. 
However, these approaches tightly couple semantic inference with trajectory prediction without an explicit structured control interface, rendering behavior-aware and constraint-aware adaptations implicit.

Among others, a key limitation lies in temporal coordination. 
End-to-end reasoning models perform full forward inference at each control step~\cite{zitkovich2023rt, shao2024lmdrive}, incurring substantial computational overhead. 
Dual-rate architectures decouple reasoning from execution: OnBoard-LLM~\cite{baumann2025enhancing} adopts intermittent large-model adjustments while preserving MPC stability; LanguageMPC~\cite{sha2023languagempc} integrates low-frequency LLM decisions with high-frequency predictive control; and 
OCP~\cite{chen2024ocp} further introduces hierarchical activation via perception confidence. 
Nevertheless, confidence-based and heuristic scheduling methods do not explicitly model behavior-level anomalies, rendering them sensitive to rapidly evolving interactions. 
Our PC-SET addresses this issue by enforcing strictly event-driven semantic activation through a lightweight gatekeeper.

Furthermore, safe semantic adaptation requires a principled control interface. Direct policy-generation methods~\cite{Zhang_2025_ICCV, shao2024lmdrive} embed reasoning directly into action outputs without separating high-level decisions from constraint-aware motion planning. LLM-guided MPC approaches~\cite{11128826, sha2023languagempc} retain kinematic safety by adjusting planner parameters within an optimization backbone. However, this reconfiguration is typically invoked without a dedicated semantic gating mechanism. In contrast, our SMPC maps discrete prompt-driven decisions to a structured predictive control interface, systematically reconfiguring the objective functions within EAMP while preserving optimization-based safety.

% --------------------------------------------------
\begin{figure*}[t!]
  \centering
  \includegraphics[width=\textwidth]{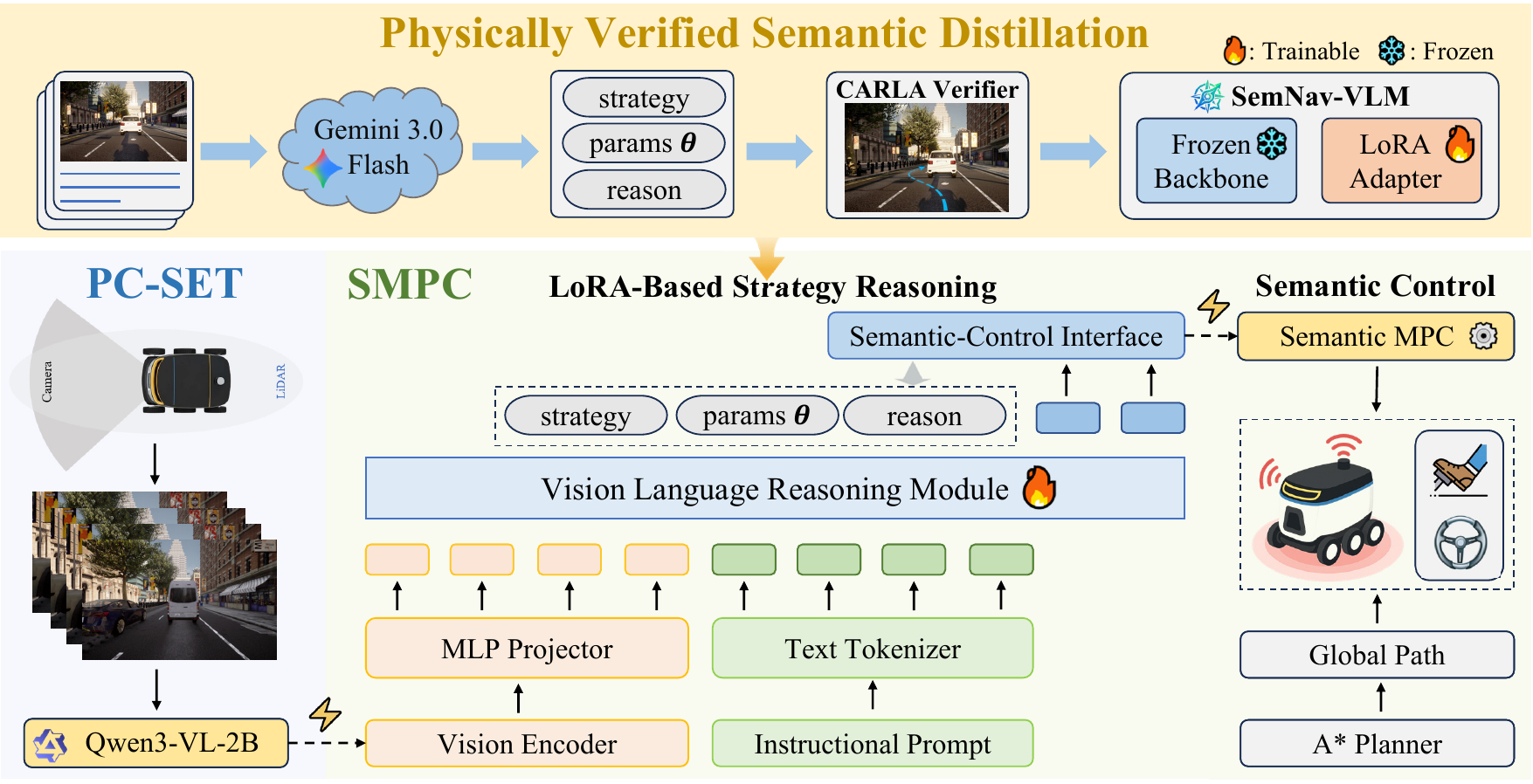}
    \caption{Overview of EAMP. Offline distillation fine-tunes SemNav-VLM. Online, PC-SET detects anomalies to trigger SMPC adaptation, decoupling reasoning from physical execution.}
  \label{fig:framework}
  \vspace{-15pt}
\end{figure*}
%--------------------------------------------------

\section{Problem Formulation and Overview of EAMP}
\label{sec:problem_formulation}

\subsection{Problem Formulation}

We consider a mobile robot operating in open and dynamic environments.
The system state at time step $t$ is denoted by
$\mathbf{x}_t = [p_t^\top, \phi_t]^\top \in \mathcal{X} \subset \mathbb{R}^3$,
where $p_t \in \mathbb{R}^2$ is the planar position and $\phi_t$ is the heading angle.
The control input is defined as $\mathbf{u}_t = [v_t, \delta_t]^\top \in \mathcal{U} \subset \mathbb{R}^2$,
consisting of the longitudinal velocity $ {v}_{t} $  and steering angle $ {\delta}_{t} $.
The system evolution follows a discrete-time kinematic model:
\begin{equation}
\mathbf{x}_{t+1} = f(\mathbf{x}_t, \mathbf{u}_t),
\label{eq:dynamics}
\end{equation}
where $f(\cdot)$ describes the nonlinear vehicle kinematics.

Local motion execution is governed by parametric MPC.
To enable diverse driving behaviors within a unified control structure,
we introduce a strategy-conditioned parameter vector
$\boldsymbol{\theta} \in \mathbb{R}^5$, defined as:
\begin{equation}
\boldsymbol{\theta}
= [w_{\mathrm{s}}, w_{\mathrm{u}}, \lambda_{\epsilon}, d_{\min}, \alpha_{\mathrm{u}}]^\top,
\label{eq:theta_def}
\end{equation}
where $w_{\mathrm{s}}$ and $w_{\mathrm{u}}$ denote weighting factors for path tracking and
control regulation, respectively,
$\lambda_{\epsilon}$ is the slack penalty weight for soft obstacle avoidance,
$d_{\min}$ specifies a nominal safety margin,
and $\alpha_{\mathrm{u}} \in [0,1]$ is a scaling factor applied to the reference control input $\mathbf{u}_k^{\mathrm{ref}}$.

Over the horizon $T$, we introduce a nonnegative slack variable $\epsilon_k \ge 0$
at each prediction step $k$ to mitigate infeasibility under tight interaction constraints.
The stage cost adopts a standard quadratic tracking formulation~\cite{10036019}:
\begin{equation}
J(\mathbf{x}_k, \mathbf{u}_k, \epsilon_k; \boldsymbol{\theta})
=
w_{\mathrm{s}} \|p_k - p_k^{\mathrm{ref}}\|^2
+ w_{\mathrm{u}} \| \mathbf{u}_k - \alpha_{\mathrm{u}} \mathbf{u}_k^{\mathrm{ref}} \|^2
+ \lambda_{\epsilon}\, \epsilon_k,
\label{eq:stage_cost}
\end{equation}
where $p_k \in \mathbb{R}^2$ encodes the predicted robot position.
The reference terms $p_k^{\mathrm{ref}}$ and $\mathbf{u}_k^{\mathrm{ref}}$
encode the nominal geometric objectives.

Then, the MPC problem computed at time step $t$ is:
\begin{subequations}\label{eq:mpc_problem}
\begin{align}
\min_{\{\mathbf{u}_k, \epsilon_k\}_{k=0}^{T-1}} \quad &
\sum_{k=0}^{T-1} J(\mathbf{x}_k, \mathbf{u}_k, \epsilon_k; \boldsymbol{\theta}),
\label{eq:mpc_obj} \\
\text{s.t.} \quad
& \mathbf{x}_{k+1} = f(\mathbf{x}_k, \mathbf{u}_k), \\
& \mathbf{u}_k \in \mathcal{U}, \quad \mathbf{x}_k \in \mathcal{X}, \quad \forall k, \\
& \mathrm{dist}(p_k, \mathcal{O}_k) \ge d_{\min} - \epsilon_k, \quad \epsilon_k \ge 0, \quad \forall k, \\
& \mathbf{x}_0 = \mathbf{x}_t,
\end{align}
\end{subequations}
where $\mathcal{O}_k$ denotes the predicted obstacle set at step $k$, 
and $\mathrm{dist}(p_k, \mathcal{O}_k)$ is the Euclidean distance from $p_k$ to $\mathcal{O}_k$.

Conventionally, the parameter vector $\boldsymbol{\theta}$ is assumed constant during execution.
However, varying interaction contexts in open environments require dynamic trade-offs
between safety margins and operational efficiency.
Consequently, the core challenge is twofold:
1) determining \emph{when} to adapt the control objective, and
2) \emph{how} to systematically translate semantic events into structured modifications
of the parameterized MPC formulation.

\subsection{Overview of EAMP Framework}

To address the above challenges, we propose EAMP, a hierarchical semantic control framework illustrated in Fig.~\ref{fig:framework}. 
Specifically, this architecture decouples semantic reasoning from continuous motion execution through a three-level design: 
1) PC-SET determines \emph{when} interaction semantics require adaptation; 
2) SMPC specifies \emph{how} strategy-level decisions are mapped to structured modifications of the optimization objective and reference; 
and 3) the underlying control layer ensures dynamically feasible motion generation under the updated configurations.

To enable reliable real-time reasoning, the strategy-level policy of SMPC is driven by a lightweight VLM, termed SemNav-VLM, obtained via physically verified semantic distillation. 
By separating semantic event detection, strategy reasoning, and optimization-based control, EAMP achieves \emph{timely}, \emph{interpretable}, and \emph{safety-preserving} adaptation in open and dynamic environments.

\section{EAMP Algorithm Design}
\label{sec:methodology}

In open environments, dynamic interaction semantics often render fixed control objectives suboptimal. To address this issue, the proposed EAMP framework employs a dual-phase architecture summarized in Algorithm~\ref{alg:esmpc_overview}. During online execution, a PC-SET determines when adaptation is required, while the SMPC dictates how to safely embed this high-level intent into continuous optimization. This real-time reasoning is empowered by an offline physically verified semantic distillation pipeline that trains a lightweight VLM. The detailed designs of EAMP are introduced below.

%--------------------------------------------------------
\begin{figure}[t] 
\vspace*{-5.5pt} 

\begin{algorithm}[H]
\caption{Overall Execution Pipeline of EAMP}
\label{alg:esmpc_overview}
\small
\begin{algorithmic}[1]
    \Statex \textbf{Initialize:} System state $\mathbf{x}_0$, nominal parameters $\boldsymbol{\theta}_{\mathrm{nom}}$, nominal path $\mathcal{P}_{\mathrm{nom}}$

\vspace{1mm}
\Statex \textbf{Phase 1: Physically Verified Semantic Distillation (Offline)}
\State Teacher strategy $s^* \leftarrow \operatorname*{arg\,max}_{s \in \mathcal{S}} p_T(s \mid \mathcal{I}, \mathbf{x}, \mathcal{P}_{\mathrm{strategy}})$ via $\mathcal{F}_T$
\State Collect dataset $\mathcal{D}_{\mathrm{pv}} \leftarrow \{ (\mathcal{I}, \mathbf{x}, s^*) \mid \text{feasible}(s^*) \}$
\State Distill student policy by minimizing $\mathbb{E}_{\mathcal{D}_{\mathrm{pv}}} [-\log \pi_{\mathrm{sem}}]$

\vspace{1mm}
\Statex \textbf{Phase 2: Event-Adaptive Control (Online)}
\For{each control step $t = 0, 1, 2, \dots$}
    \State \textit{\# Module A: Prompt-Configurable Semantic Event Trigger}
    \State Construct temporal clip $\mathbf{I}_c = (\mathcal{I}_{c}, \mathcal{I}_{c-1}, \mathcal{I}_{c-2})$
    \State Evaluate visual anomaly to obtain stable trigger $\tau_t^* \in \{0, 1\}$
    
    \State \textit{\# Module B: Semantic Model Predictive Control}
    \State Evaluate sustained activation flag $\sigma_t \in \{0, 1\}$
    \If{$\sigma_t = 1$}
        \State $s_t \leftarrow \operatorname*{arg\,max}_{s \in \mathcal{S}} \pi_{\mathrm{sem}}(s \mid \mathcal{I}_t, \mathbf{x}_t, \mathcal{P}_{\mathrm{strategy}})$
        \State Adapt configuration: $\boldsymbol{\theta}_t \leftarrow \Gamma(s_t)$, $\mathcal{P}_t \leftarrow \Psi(s_t, \mathcal{P}_{\mathrm{nom}})$
    \Else
        \State Maintain nominal: $\boldsymbol{\theta}_t \leftarrow \boldsymbol{\theta}_{\mathrm{nom}}$, $\mathcal{P}_t \leftarrow \mathcal{P}_{\mathrm{nom}}$
    \EndIf
    \State Solve MPC optimization with $(\boldsymbol{\theta}_t, \mathcal{P}_t)$ to obtain control $\mathbf{u}_t$
    \State Apply control input $\mathbf{u}_t$ to the vehicle
\EndFor
\end{algorithmic}
\end{algorithm}
 
\vspace*{-15pt}
\end{figure}
%------------------------------------------------------------

\subsection{Prompt-Configurable Semantic Event Trigger}
\label{sec:pc_set}

\begin{figure}[t]
    \vspace{3.5pt}
    \centering
    \begin{subfigure}[t]{0.48\columnwidth}
        \centering
        \includegraphics[width=\linewidth]{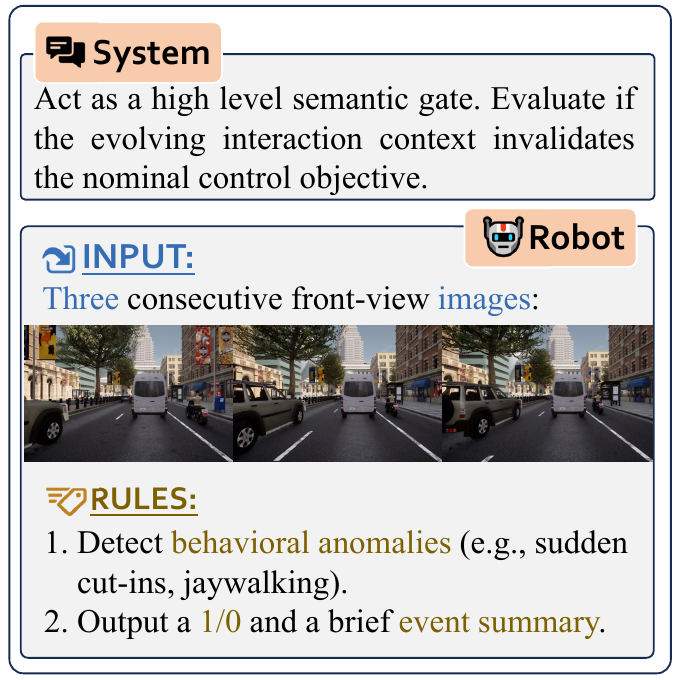}
        \caption{PC-SET prompt.}
        \label{fig:prompt_gate}
    \end{subfigure}
    \hfill
    \begin{subfigure}[t]{0.48\columnwidth}
        \centering
        \includegraphics[width=\linewidth]{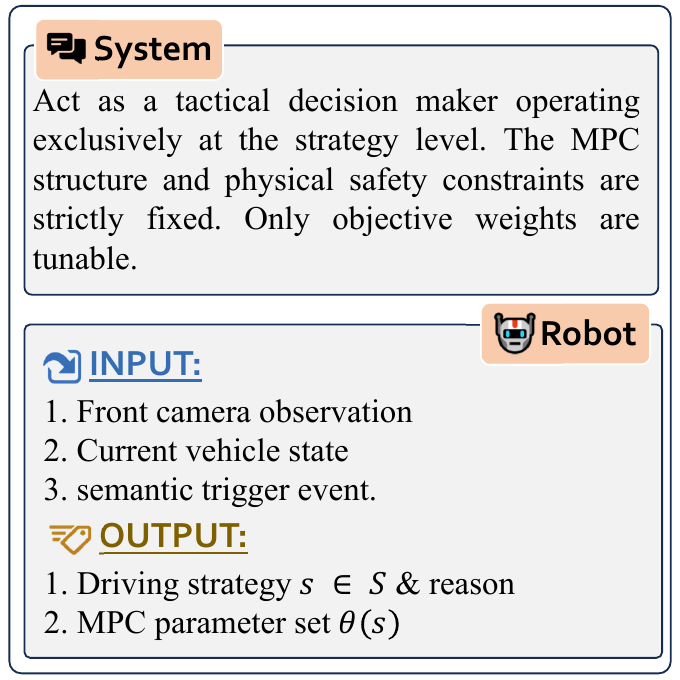}
        \caption{SMPC prompt.}
        \label{fig:prompt_strategy}
    \end{subfigure}
    \caption{Hierarchical prompt design. \emph{Left}: PC-SET for event detection. \emph{Right}: SMPC for strategy-level MPC adaptation.}
    \label{fig:prompt_design}
    \vspace{-10pt}
\end{figure}

To determine \emph{when} to adapt the nominal MPC objective,
we introduce PC-SET, a behavior-level semantic event trigger
running asynchronously at up to 3\,Hz using a VLM (Qwen3-VL-2B-Instruct).
As shown in Fig.~\ref{fig:prompt_design}(a), instead of merely classifying objects,
PC-SET evaluates short temporal clips for dynamic anomalies and outputs a discrete binary signal $\tau_c^*$ to activate event-adaptive semantic control,
leaving continuous motion optimization to the underlying MPC.

\textbf{Prompt-conditioned semantic assessment.}
To capture short-term interaction dynamics, the input is constructed as
a temporal clip of the three most recent frames,
denoted as $\mathbf{I}_c = (\mathcal{I}_{c}, \mathcal{I}_{c-1}, \mathcal{I}_{c-2})$, where $c$ denotes the discrete step of the asynchronous semantic evaluation.
Given the visual clip $\mathbf{I}_c$ and a semantic prompt $\mathcal{P}_{\mathrm{trigger}}$, a lightweight VLM $\mathcal{F}_{\mathrm{gate}}$ acts as a continuous semantic monitor. 
Instead of interpreting ambiguous text or computing continuous probabilities, the model is configured to explicitly output a binary anomaly indicator alongside a descriptive text $y_c$ solely when an anomaly is present:
\begin{equation}
    \mathcal{F}_{\mathrm{gate}}(\mathbf{I}_c, \mathcal{P}_{\mathrm{trigger}}) = 
    \begin{cases}
        (1, y_c), & \text{if anomaly detected}, \\
        (0, \emptyset), & \text{otherwise}.
    \end{cases}
    \label{eq:vlm_output}
\end{equation}

\textbf{Decision triggering via temporal consistency.}
Since semantic interpretation may be affected by transient visual
noise or occlusions, the instantaneous binary indicator is first extracted from the VLM output:
\begin{equation}
    \tau_c^{\mathrm{raw}} = \pi_1\left(\mathcal{F}_{\mathrm{gate}}(\mathbf{I}_c, \mathcal{P}_{\mathrm{trigger}})\right) \in \{0, 1\},
    \label{eq:raw_trigger}
\end{equation}
where $\pi_1$ projects the output tuple onto its first component.
To ensure trigger stability without introducing excessive delay, the final event-adaptive trigger signal $\tau_c^*$ enforces temporal consistency over a sliding window of length~$K$:
\begin{equation}
    \tau_c^* = \mathbb{I} (\textstyle \sum_{k=0}^{K-1} \tau_{c-k}^{\mathrm{raw}} \ge m ),
    \label{eq:temporal_filter}
\end{equation}
where $m \le K$ is an activation threshold and $\mathbb{I}(\cdot)$ is the indicator function.
This discrete signal $\tau_c^*$ is held constant via a zero-order hold (ZOH) and resampled at the high-frequency MPC time step $t$, producing $\tau_t^*$ to activate either the nominal or event-adapted MPC configuration. 
This induces a hybrid structure: semantic decisions evolve on an event-adaptive timescale, while continuous motion optimization remains governed by the MPC.

%--------------------------------------------------
%--------------------------------------------------

\subsection{Semantic Model Predictive Control}
\label{sec:smpc}

Driven by the event-triggered signal $\tau_t^*$ in the EAMP framework,
the SMPC module addresses
\emph{how} high-level semantic intent is embedded into a continuous and
constraint-aware MPC formulation.
Rather than generating control inputs or trajectories directly,
semantic reasoning operates at the strategy level and selects among
predefined MPC configuration templates.
This design enables structured, event-driven reconfiguration of MPC
objectives and references while preserving feasibility, stability, and
real-time solvability.

\textbf{Strategy-level semantic decision.}
To abstract semantic intent into a compact and interpretable form, we
discretize the decision space into a finite set of interaction strategies
$\mathcal{S}$ (e.g., yielding, cautious following, overtaking), without specifying low-level control actions.
Upon event activation (i.e., $\tau_t^* = 1$), the vision-language policy
$\pi_{\mathrm{sem}}$ (instantiated by SemNav-VLM)
infers the most plausible strategy based on the
current visual context and ego state. To ensure safety during inference latency, the system maintains its prior control configuration until reasoning completes, locking the result as the active strategy $s_t$:
\begin{equation}
    s_t = \operatorname*{arg\,max}_{s \in \mathcal{S}}
    \ \pi_{\mathrm{sem}}(s \mid \mathcal{I}_t, \mathbf{x}_t,
    \mathcal{P}_{\mathrm{strategy}}).
    \label{eq:action_generation}
\end{equation}
This formulation decouples logic-level semantic reasoning from numerical
optimization by mapping complex interaction understanding into a
discrete decision variable.
Importantly, restricting semantic outputs to
$s_t \in \mathcal{S}$ yields a verifiable interface to MPC, preventing
unstructured semantic predictions from directly perturbing the optimizer.

\textbf{Template-based MPC reconfiguration.}
To systematically embed the inferred strategy into the MPC formulation,
we define a \emph{control configuration space}
$\Omega = \Theta \times \mathbb{P}$,
where $\Theta$ denotes the admissible parameter set and
$\mathbb{P}$ denotes the feasible reference path space.
Each $s \in \mathcal{S}$ is uniquely mapped to a predefined, scenario-agnostic MPC configuration template that defines a specific level of control conservativeness, enabling reproducible semantic adaptation.

Given the active strategy $s_t$, SMPC constructs the control
configuration through two complementary mappings.
First, the objective and constraint parameters are selected via the
template mapping function $\Gamma$:
\begin{equation}
    \tilde{\boldsymbol{\theta}}_t = \Gamma(s_t)
    \triangleq
    [w_{\mathrm{s}}(s_t),\ w_{\mathrm{u}}(s_t),\ \lambda_{\epsilon}(s_t),\
    d_{\min}(s_t),\ \alpha_{\mathrm{u}}(s_t)]^\top.
    \label{eq:theta_mapping}
\end{equation}
Second, the geometric intent of the strategy is reflected by adapting
the global plan through the reference mapping function $\Psi$:
\begin{equation}
    \tilde{\mathcal{P}}_t = \Psi(s_t, \mathcal{P}_{\mathrm{nom}}).
    \label{eq:path_mapping}
\end{equation}
Together, $(\tilde{\boldsymbol{\theta}}_t, \tilde{\mathcal{P}}_t)$ specifies
a structured update to the MPC problem without
altering its core optimization architecture.

\textbf{Event-adaptive switching and stability.}
To suppress chattering and ensure consistent execution of dynamic maneuvers, we define a sustained activation flag $\sigma_t \in \{0, 1\}$, which remains active if $\tau_t^* = 1$ or if the system is within a minimum hold duration since the last trigger.
The active MPC configuration is then governed by the event-triggered update law:
\begin{equation}
    (\boldsymbol{\theta}_t, \mathcal{P}_t) =
    \begin{cases}
        (\tilde{\boldsymbol{\theta}}_t, \tilde{\mathcal{P}}_t), & \sigma_t = 1, \\
        (\boldsymbol{\theta}_{\mathrm{nom}}, \mathcal{P}_{\mathrm{nom}}),
        & \sigma_t = 0.
    \end{cases}
    \label{eq:hybrid_update}
\end{equation}
As a result, semantic adaptation evolves on an event-driven timescale,
while continuous motion optimization remains entirely within the MPC loop.

\subsection{Physically Verified Semantic Distillation}
\label{sec:distillation}

The SMPC framework requires a strategy-level semantic policy
$\pi_{\mathrm{sem}}$ that maps a visual observation and ego state
to a discrete interaction strategy $s \in \mathcal{S}$,
which subsequently conditions MPC reconfiguration.
However, interaction strategies lack explicit ground-truth annotations,
making direct supervised learning infeasible.

To address this issue, we adopt a teacher--student distillation paradigm.
A large cloud-based VLM (Gemini-3.0-Flash) serves as the semantic teacher,
producing strategy hypotheses under the structured prompt shown in Fig.~\ref{fig:prompt_design}(b).
Instead of treating teacher outputs as direct supervision,
we introduce a physical verification stage to ensure consistency
between semantic plausibility and dynamic feasibility.

Specifically, candidate strategies are validated through
closed-loop evaluation under the SMPC controller in CARLA.
Only physically feasible strategies are retained to construct
a supervision set:
\begin{equation}
\mathcal{D}_{\mathrm{pv}} =
\left\{
(\mathcal{I}, \mathbf{x}, s) \;\middle|\;
s \in \mathcal{S}_{\mathrm{feasible}}(\mathcal{I}, \mathbf{x})
\right\},
\end{equation}
which filters out semantically reasonable yet dynamically infeasible
decisions.

The student semantic policy, referred to as SemNav-VLM,
is instantiated by fine-tuning Qwen3-VL-8B-Instruct with LoRA~\cite{hu2022lora}.
The distillation objective minimizes the negative log-likelihood
over the physically verified dataset:
\begin{equation}
\mathcal{L}_{\mathrm{distill}}
=
\mathbb{E}_{(\mathcal{I}, \mathbf{x}, s)\sim\mathcal{D}_{\mathrm{pv}}}
\left[
- \log \pi_{\mathrm{sem}}(s \mid \mathcal{I}, \mathbf{x}, \mathcal{P}_{\mathrm{strategy}})
\right].
\end{equation}
The complete distillation pipeline is detailed in
Algorithm~\ref{alg:pv_distill}.

%---------------------------------------------------------
\begin{figure}[t] 
\vspace*{-5.5pt} 

\begin{algorithm}[H]
\caption{Physically Verified Semantic Distillation}
\label{alg:pv_distill}
\small
\begin{algorithmic}[1]
\Statex \textbf{Input:} CARLA simulator $\mathcal{E}$, strategy set $\mathcal{S}$, teacher VLM $\mathcal{F}_T$, student VLM $\mathcal{F}_S$, SMPC controller $\mathcal{C}_{\mathrm{SMPC}}$
\Statex \textbf{Output:} Distilled semantic policy $\pi_{\mathrm{sem}}$
\State $\mathcal{D}_{\mathrm{pv}} \leftarrow \emptyset$
\For{each interaction episode in $\mathcal{E}$}
    \State collect observation $\mathcal{I}$ and ego state $\mathbf{x}$
    \State $s^* \leftarrow \operatorname*{arg\,max}_{s \in \mathcal{S}} p_T(s\mid\mathcal{I},\mathbf{x}, \mathcal{P}_{\mathrm{strategy}})$ via $\mathcal{F}_T$
    \State rollout closed-loop $\mathcal{C}_{\mathrm{SMPC}}$ with strategy $s^*$ in $\mathcal{E}$
    \If{$s^*$ is physically feasible (i.e., collision-free)}
        \State $\mathcal{D}_{\mathrm{pv}} \leftarrow \mathcal{D}_{\mathrm{pv}} \cup \{(\mathcal{I},\mathbf{x},s^*)\}$
    \EndIf
\EndFor
\State initialize LoRA parameters on $\mathcal{F}_S$
\While{validation loss not converged}
    \State sample a verified batch $(\mathcal{I},\mathbf{x},s) \sim \mathcal{D}_{\mathrm{pv}}$
    \State update $\pi_{\mathrm{sem}}$ by minimizing $-\log \pi_{\mathrm{sem}}(s\mid\mathcal{I},\mathbf{x}, \mathcal{P}_{\mathrm{strategy}})$
\EndWhile
\State \Return $\pi_{\mathrm{sem}}$
\end{algorithmic}
\end{algorithm}

\vspace*{-15pt}
\end{figure}
%-------------------------------------------------------------

\section{Experiments}

\subsection{Experimental Setup}
\label{sec:setup}

Experiments are conducted in CARLA utilizing ROS for closed-loop integration of perception, reasoning, and control. We employ Town06 and Town10 to simulate highway and dense urban environments. The ego vehicle is a delivery robot equipped with a front-view RGB camera and a 64-line LiDAR, exchanging data via the Carla-ROS-bridge to enable real-time system interaction.

A local MPC controller governs motion execution at 20\,Hz, using a prediction horizon $T=18$ and time step $\Delta t=0.3$\,s. To ensure fair geometric comparisons, all methods track a shared A$^*$-generated global route. All evaluations are performed using two NVIDIA RTX 3090 GPUs. For comparison, we implement the following baselines:
1) \textbf{RDA}~\cite{10036019}: A geometric MPC planner without semantic reasoning;
2) \textbf{PCS}~\cite{tanwani2020rilaas}: A periodical collaboration strategy that queries the VLM at fixed intervals;
3) \textbf{OCP}~\cite{chen2024ocp}: An opportunistic collaboration scheme triggered by unknown object perception confidence.

System performance is evaluated using five metrics:
1) success rate ($\mathrm{SRate}$), measuring the ratio of collision-free and timeout-free trials;
2) minimum time-to-collision ($\mathrm{TTC}$), indicating the worst-case dynamic safety margin;
3) finishing time ($\mathrm{FTime}$), reflecting navigation efficiency;
4) trajectory length ($\mathrm{TLen}$), representing spatial efficiency; and
5) speed variability ($\mathrm{SVar}$), characterizing control smoothness.

\subsection{Experiment 1: Evaluation of SemNav-VLM}
\label{sec:exp1}

The semantic strategy reasoning capability of SemNav-VLM is evaluated against representative baselines. To fine-tune and rigorously test the model, we extract interactive keyframes from continuous driving episodes within the CARLA Town04 and Town10 environments. This yields a behavior-annotated dataset comprising 2000 training samples for LoRA fine-tuning and a 200-sample held-out set for offline evaluation. Each snapshot pairs a front-view RGB image with the concurrent ego-vehicle state. Performance is measured via strategy accuracy ($\mathrm{SAcc}$), where a prediction is deemed correct strictly if it matches the physically verified label obtained through closed-loop continuous rollout. This physical verification protocol ensures that all reference strategies are dynamically feasible under complex interactions.

%---------------------- Experiment1 ----------------------------
\begin{table}[t]
\vspace{3.5pt}
\centering
\caption{Comparison of inference latency and SAcc.}
\label{tab:distillation_metrics}
\resizebox{\columnwidth}{!}{
\begin{tabular}{lccc}
\toprule
\textbf{Model} & \textbf{Deployment} & \textbf{Latency (ms) $\downarrow$ } & \textbf{SAcc (\%) $\uparrow$} \\
\midrule
Gemini-3.0-Flash        & Cloud API & 11569 & 86 \\
Qwen3-VL-8B-Instruct    & Local     & 2117  & 42 \\
\rowcolor{blue!8}
SemNav-VLM              & \textbf{Local} & \textbf{1701} & \textbf{96} \\
\bottomrule
\end{tabular}}
\end{table}
%--------------------------------------------------

%--------------------------------------------------
\begin{figure}[t]
	\centering
	\noindent
	\includegraphics[width=0.48\columnwidth]{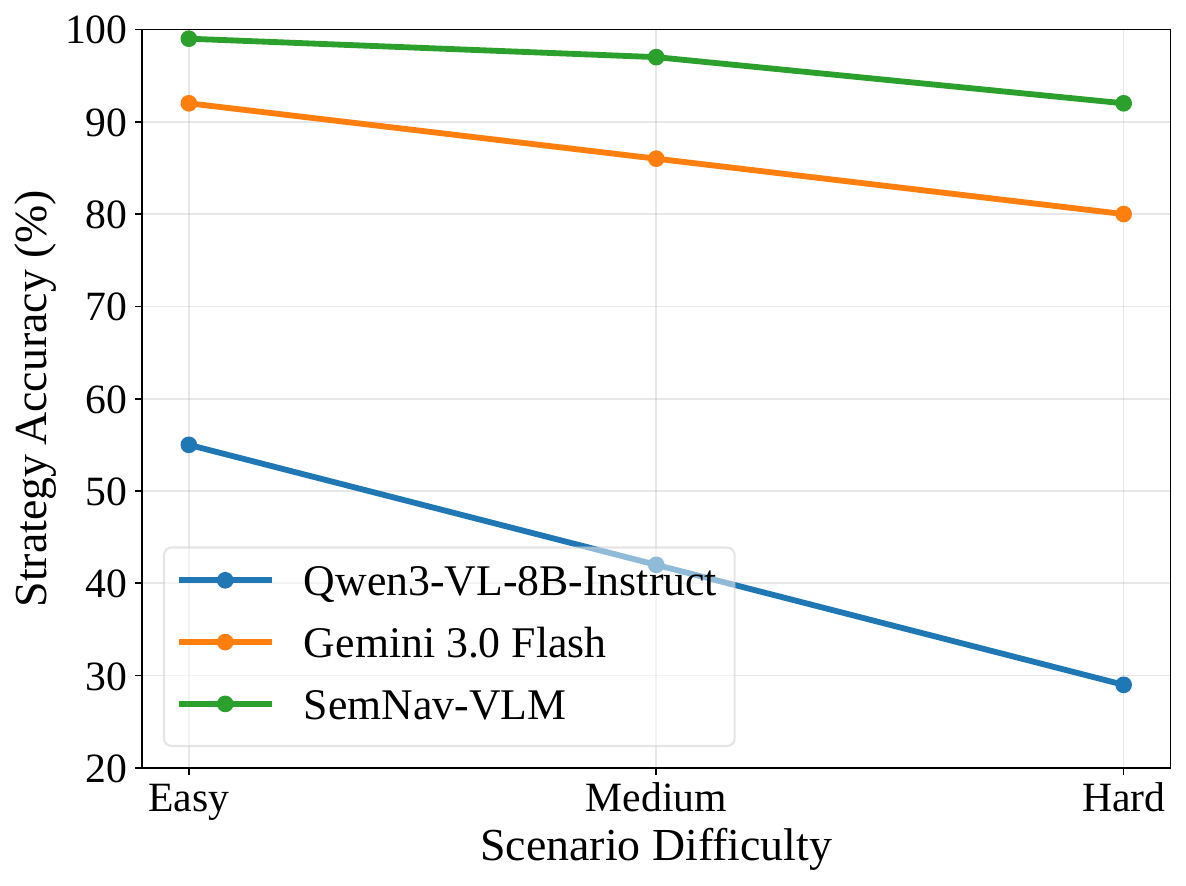}%
	\includegraphics[width=0.48\columnwidth]{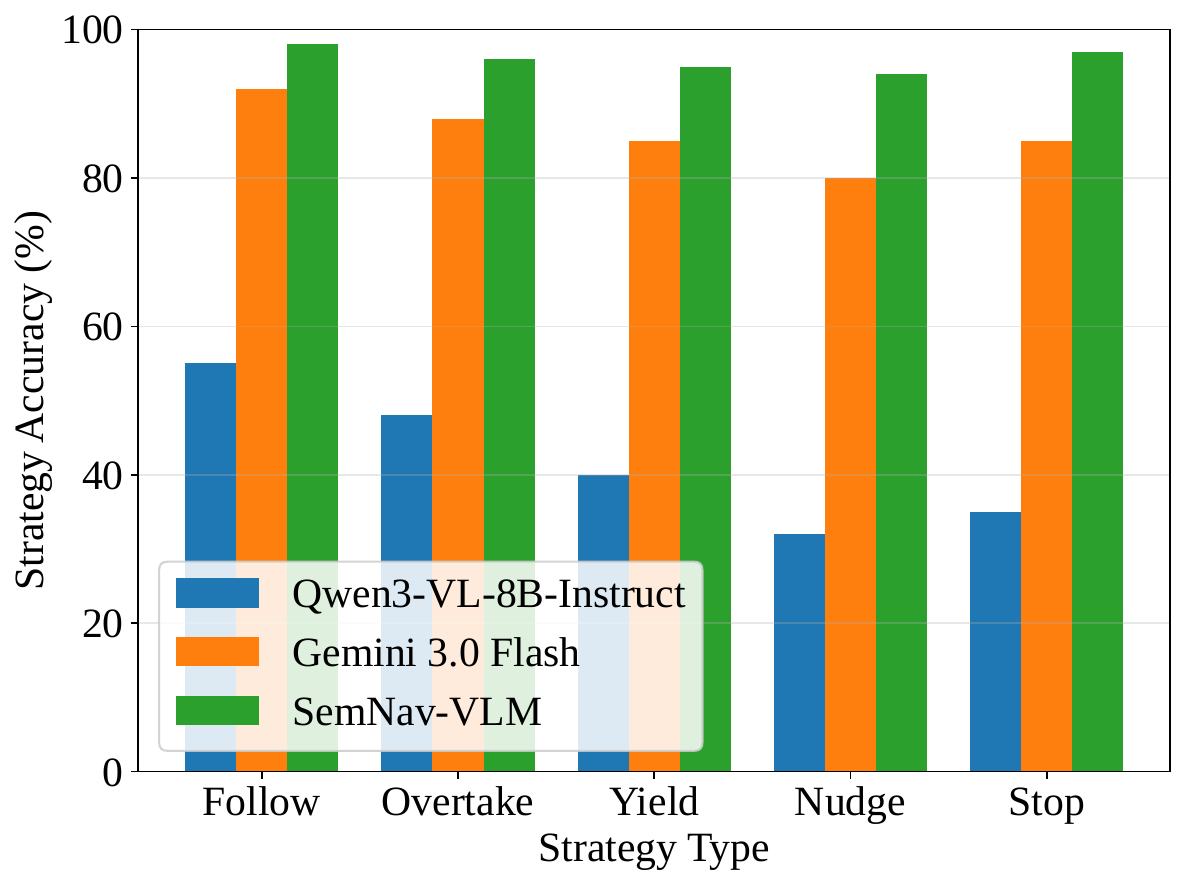}%
	\caption{Strategy prediction performance across scenario difficulties (\emph{left}) and specific maneuvers (\emph{right}).}
	\label{fig:exp1_quant}
    \vspace{-10pt}
\end{figure}
%--------------------------------------------------

%--------------------------------------------------
\begin{figure}[t]
    \vspace{3.5pt}
    \centering
    \includegraphics[width=\linewidth]{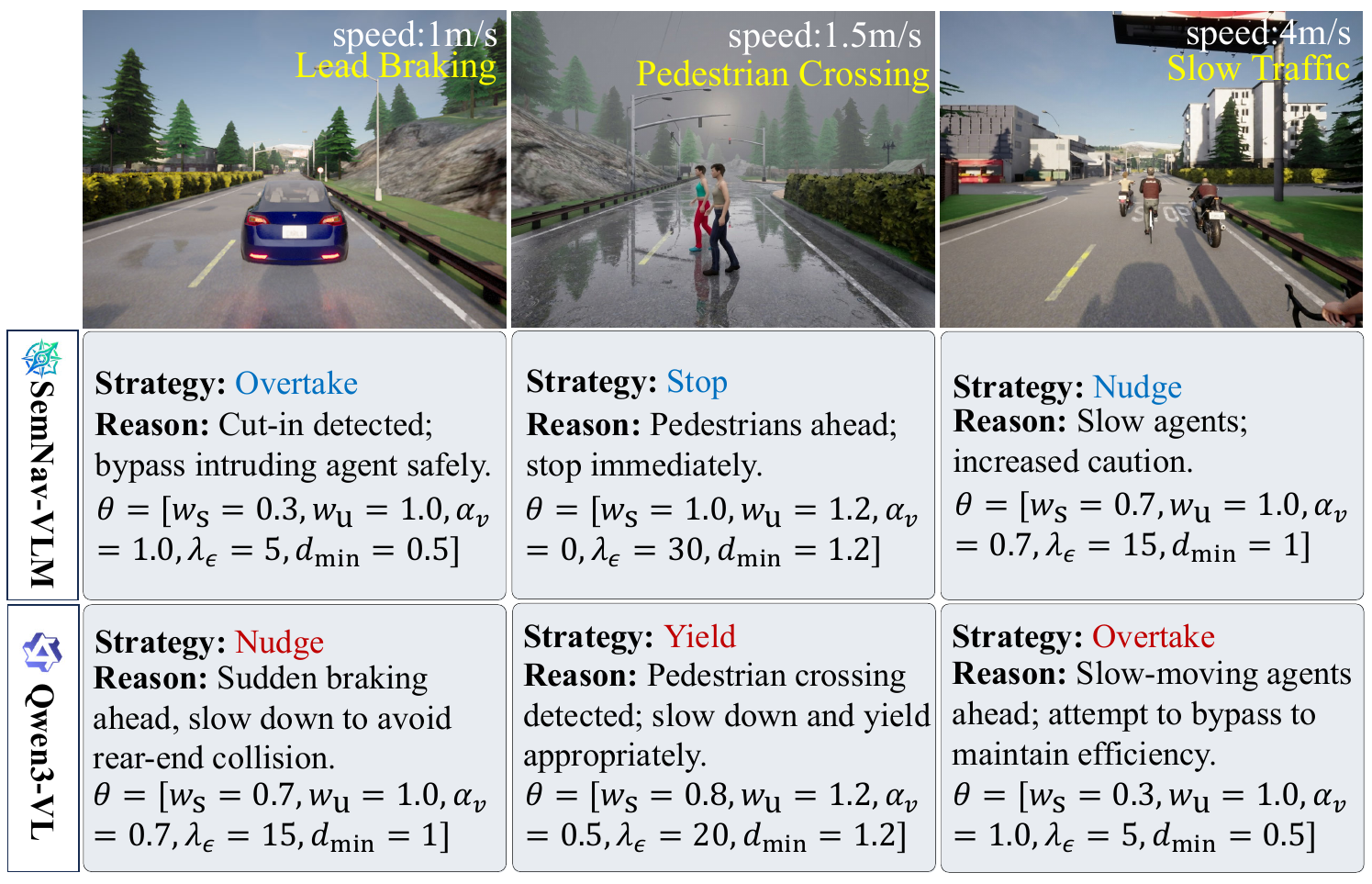}
    \caption{Qualitative strategy inference and adaptation under three typical semantic events. \emph{Top}: SemNav-VLM. \emph{Bottom}: Qwen3-VL-8B-Instruct.}
    \label{fig:exp1_qual}
    \vspace{-10pt}
\end{figure}
%--------------------------------------------------

\textbf{Quantitative results.} 
Table~\ref{tab:distillation_metrics} summarizes the overall performance on the held-out evaluation set. SemNav-VLM achieves a 96\% strategy accuracy, establishing a substantial margin over the local base model Qwen3-VL-8B-Instruct~\cite{bai2025qwen3} (42\%) and securely surpassing the zero-shot cloud-based teacher Gemini-3.0-Flash (86\%). As illustrated in Fig.~\ref{fig:exp1_quant}, SemNav-VLM maintains robust accuracy across all difficulty levels, whereas the base model suffers a sharp degradation in interaction-heavy scenarios. Furthermore, the per-strategy breakdown highlights that the proposed physically verified distillation particularly enhances reasoning about safety-critical maneuvers, such as yield, nudge, and stop.

\textbf{Qualitative results.} 
Fig.~\ref{fig:exp1_qual} visualizes the closed-loop execution under three representative semantic events. In the lead vehicle braking scenario, SemNav-VLM proactively identifies a cut-in and adopts an overtake strategy to safely bypass the intruding agent, thereby maintaining navigation efficiency. In contrast, the general base model misjudges it as a generic braking event and conservatively proposes a nudge, which would unnecessarily slow down the progression. When encountering a pedestrian crossing, our model correctly triggers an immediate stop prioritizing absolute safety, rather than proposing a hesitant yield. Similarly, amidst dense slow traffic, SemNav-VLM selects a controlled nudge to balance safety and efficiency, entirely avoiding the unsafe and aggressive overtaking actions predicted by the generic baseline. These results confirm that SemNav-VLM effectively translates scene semantics into correct and executable actions for the downstream predictive controller.

%---------------------D Experiment2 ----------------------------
\begin{figure*}[t]
    \centering
    \begin{subfigure}[t]{0.32\textwidth}
        \centering
        \includegraphics[width=\linewidth]{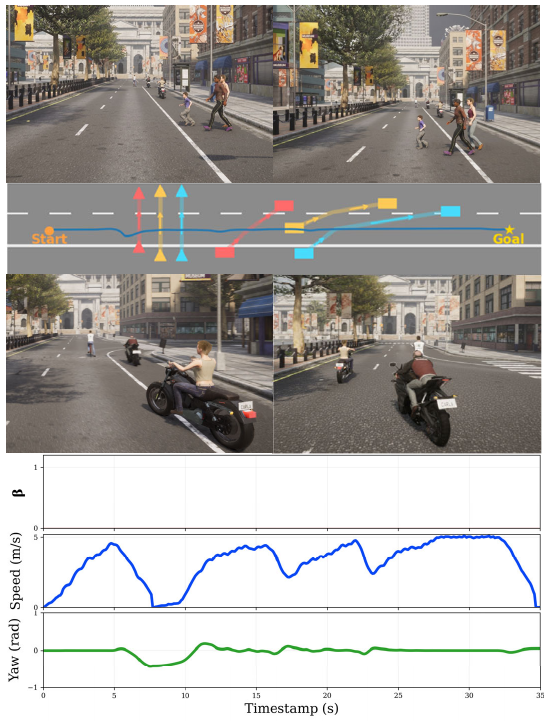}
        \caption{RDA}
        \label{fig:exp3_mpc}
    \end{subfigure}\hfill
    \begin{subfigure}[t]{0.32\textwidth}
        \centering
        \includegraphics[width=\linewidth]{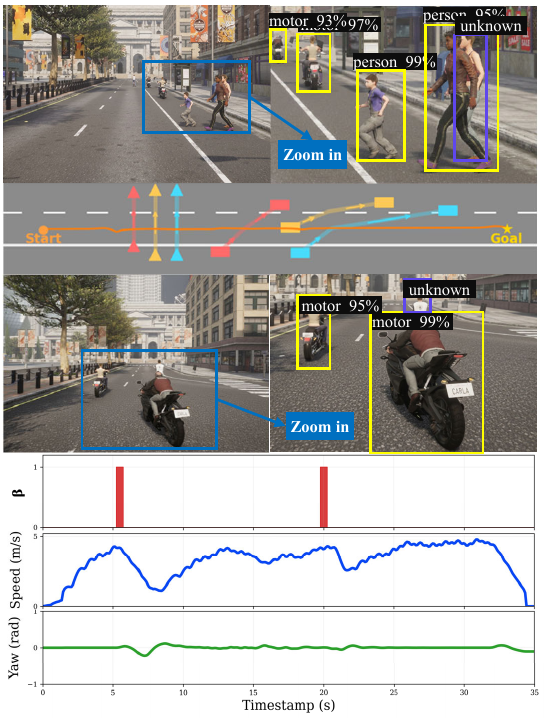}
        \caption{OCP}
        \label{fig:exp3_ocp}
    \end{subfigure}\hfill
    \begin{subfigure}[t]{0.32\textwidth}
        \centering
        \includegraphics[width=\linewidth]{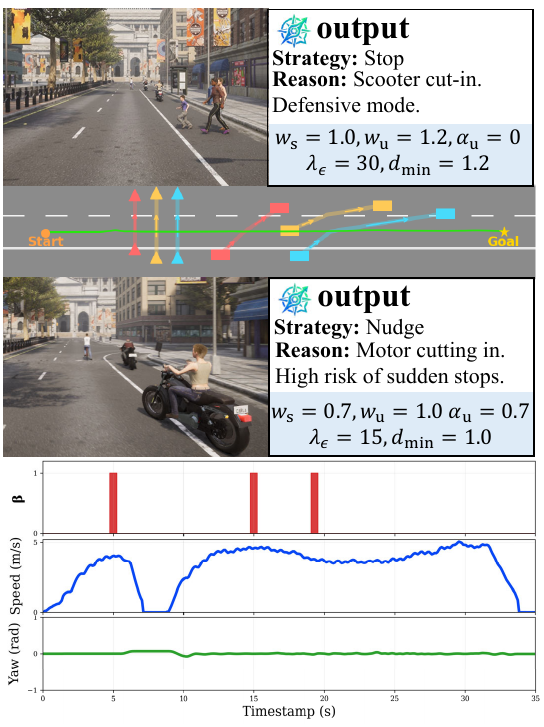}
        \caption{EAMP (Ours)}
        \label{fig:exp3_esmcp}
    \end{subfigure}

    \caption{Qualitative comparison of closed-loop executions facing sequential pedestrian jaywalking and sudden cut-ins.}

    \label{fig:exp3_compare}
    \vspace{-10pt}
\end{figure*}
%-------------------------------------------------

\subsection{Experiment 2: Evaluation of EAMP}
\label{sec:exp_smpc}

We evaluate the proposed EAMP framework within a custom-built smart logistics scenario in the CARLA Town10 map, which features dense urban geometry, frequent occlusions, and heterogeneous traffic. To systematically examine scalability against interaction complexity, we design easy, medium, and hard difficulty levels. These levels incrementally introduce challenging interactions commonly encountered in last-mile delivery tasks, including unsignalized pedestrian crossings and sudden two-wheeler cut-ins. All methods are evaluated under identical global routes and twenty trials with randomized agent behaviors.

\textbf{Quantitative results.}
Table~\ref{tab:ea_mpc_results} reports the quantitative performance averaged across twenty independent trials. EAMP demonstrates a superior safety margin during critical interaction moments. Specifically in the hard scenario, EAMP achieves a $\mathrm{TTC}$ of 1.77\,s, representing a 32\% improvement over OCP and a 30\% improvement over PCS. This substantial safety advantage exposes the core limitations of existing coordination schemes: PCS invokes the VLM based on fixed intervals, inherently missing the sudden onset of hazards; meanwhile, OCP relies on perception confidence drops, struggling to identify dangerous behavioral shifts of well-tracked, known entities. By employing PC-SET to continuously evaluate behavioral anomalies, EAMP guarantees timely semantic adaptation. Furthermore, this proactive capability does not compromise navigation efficiency. Under hard conditions, EAMP achieves the fastest completion time of 34.2\,s. Crucially, this efficiency is attained alongside strictly bounded speed variability. EAMP maintains an $\mathrm{SVar}$ of 1.40\,m/s, ensuring significantly smoother control than PCS (1.50\,m/s) and OCP (1.45\,m/s) by avoiding the periodic parameter jumps and delayed reactive corrections inherent to the baselines. Similar advantages are consistently observed in the easy and medium scenarios, indicating that the proposed event-triggered semantic adaptation maintains robust safety efficiency trade-offs across varying interaction complexities.

%--------------------------------------------------
\begin{table}[t]
\centering
\caption{Quantitative comparison across three difficulty levels in Town10 (20 trials).}
\label{tab:ea_mpc_results}
\small
\renewcommand{\arraystretch}{1.15}

% Highlight the proposed method without implying that every value is optimal.
\newcommand{\ours}[1]{\cellcolor{blue!8}#1}

\begin{adjustbox}{max width=\linewidth}
\begin{tabular}{l l c c c c}
\toprule
\textbf{Level} & \textbf{Method} &
$\textbf{TTC}$ (s) $\uparrow$ & \textbf{FTime} (s) $\downarrow$ & \textbf{TLen} (m) $\downarrow$ & \textbf{SVar} (m/s) $\downarrow$ \\
\midrule

\multirow{4}{*}{Easy}
& RDA     & 1.30 & 31.3 & 113.0 & 1.56 \\
& OCP     & 1.33 & \textbf{30.9} & 112.8 & 1.53 \\
& PCS     & 1.38 & 31.2 & 113.1 & 1.60 \\
& \ours{EAMP} & \ours{\textbf{2.13}} & \ours{32.8} & \ours{\textbf{112.5}} & \ours{\textbf{1.38}} \\
\midrule

\multirow{4}{*}{Medium}
& RDA     & 1.22 & \textbf{33.4} & 113.7 & 1.58 \\
& OCP     & 1.45 & 33.9 & 112.8 & 1.49 \\
& PCS     & 1.48 & 34.2 & 113.0 & 1.54 \\
& \ours{EAMP} & \ours{\textbf{1.86}} & \ours{34.1} & \ours{\textbf{112.3}} & \ours{\textbf{1.39}} \\
\midrule

\multirow{4}{*}{Hard}
& RDA     & 1.13 & 34.5 & 113.8 & 1.57 \\
& OCP     & 1.34 & 34.3 & 112.6 & 1.45 \\
& PCS     & 1.36 & 34.6 & 112.9 & 1.50 \\
& \ours{EAMP} & \ours{\textbf{1.77}} & \ours{\textbf{34.2}} & \ours{\textbf{112.3}} & \ours{\textbf{1.40}} \\
\bottomrule
\end{tabular}
\end{adjustbox}
\vspace{-10pt}
\end{table}
%--------------------------------------------------

\textbf{Qualitative comparison.}
To intuitively understand the source of these quantitative improvements, Fig.~\ref{fig:exp3_compare} illustrates representative closed-loop executions from a hard scenario, which features sequential pedestrian jaywalking followed by multiple two-wheeler cut-ins. 
The purely geometric RDA reacts based solely on instantaneous spatial constraints, delaying yielding maneuvers until pedestrians or cutting-in vehicles breach its immediate safety region, resulting in abrupt emergency braking. 
OCP triggers semantic intervention based on perception confidence, but its reliance on object tracking reveals a critical limitation: when already-recognized entities (e.g., visible pedestrians) suddenly change their behaviors and cross the road, the perception confidence remains high, causing the semantic intervention to arrive dangerously late. 
In contrast, EAMP explicitly targets behavioral anomalies via PC-SET. Upon detecting the unexpected jaywalking or the rapid succession of cut-ins, the SMPC module instantly reconfigures the controller. This precisely timed adaptation enables early yielding and proactive defensive following, translating into visibly smoother deceleration curves and confirming a safer, anticipatory motion execution.

\subsection{Experiment 3: Semantic Triggering Analysis}
\label{sec:exp_c}

\begin{figure}[t]
    \centering
    \includegraphics[width=\linewidth]{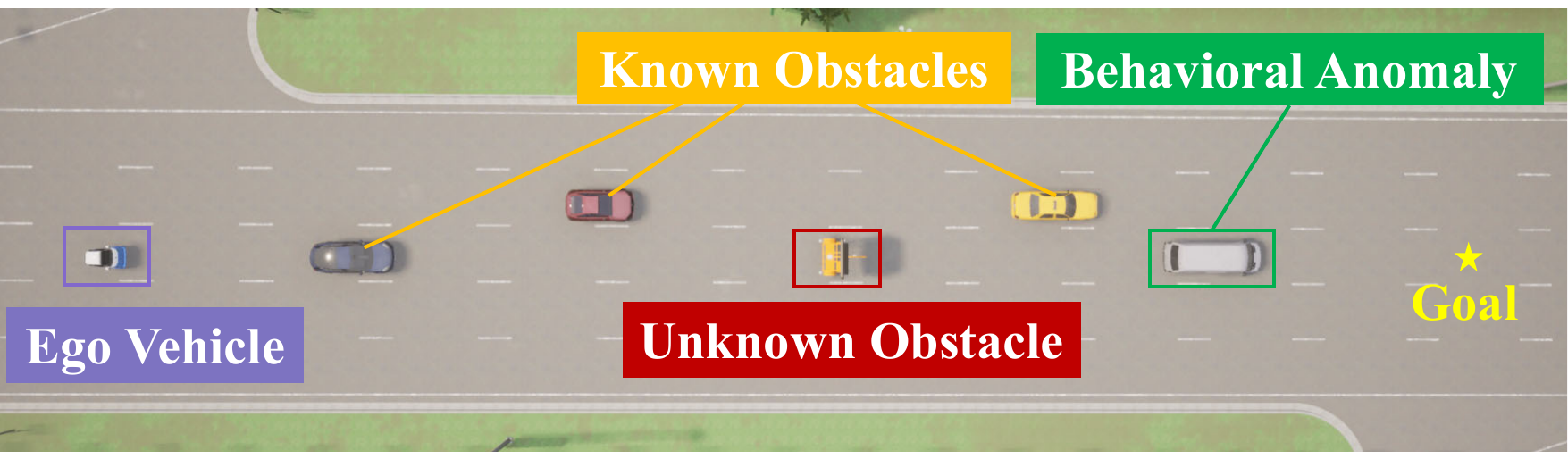}
    \caption{CARLA Town06 scenario for Experiment 3.}
    \label{fig:exp_c_setup}
    \vspace{-10pt}
\end{figure}

This experiment isolates the temporal performance of semantic triggering under dynamic behavioral anomalies where perception confidence remains structurally stable. To rigorously evaluate against OCP, we directly extend its original experimental scenario. As illustrated in Fig.~\ref{fig:exp_c_setup}, evaluations occur along a 110\,m segment in Town06. Building upon the original OCP setup featuring regular traffic and an unknown static obstacle, we introduce a critical dynamic challenge: a recognized leading vehicle that drives normally before abruptly braking to a complete stop. We conduct twenty trials with randomized initial conditions, requiring collision-free and stagnation-free route completion.

%--------------------- Experiment3 -----------------------------
%--------------------------------------------------

\textbf{Quantitative Results.}
Table~\ref{tab:exp_c_metrics} summarizes the performance across twenty trials. EAMP achieves a 95\% $\mathrm{SRate}$ compared to OCP's 60\%. This contrast empirically exposes the vulnerability of confidence-based triggering. While OCP handles its original static obstacle adequately, it fails fundamentally when a thoroughly tracked entity exhibits a sudden behavioral shift. Despite executing proactive maneuvers against this new anomaly, EAMP concurrently reduces the average $\mathrm{FTime}$ to 37.85\,s. This efficiency is matched by a shorter $\mathrm{TLen}$ of 113.29\,m and a lower $\mathrm{SVar}$ of 1.32\,m/s, proving that PC-SET secures a superior safety envelope without conservative detours.

\begin{figure}[t]
    \centering
    \begin{subfigure}[t]{0.48\columnwidth}
        \centering
        \includegraphics[width=\linewidth]{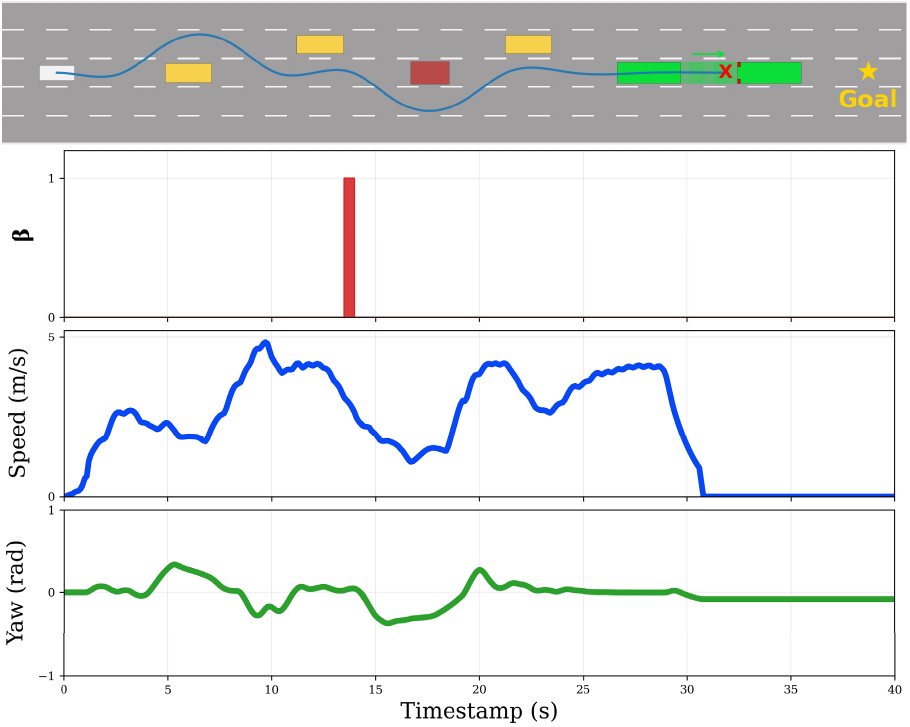}
        \caption{OCP}
    \end{subfigure}
    \hfill
    \begin{subfigure}[t]{0.48\columnwidth}
        \centering
        \includegraphics[width=\linewidth]{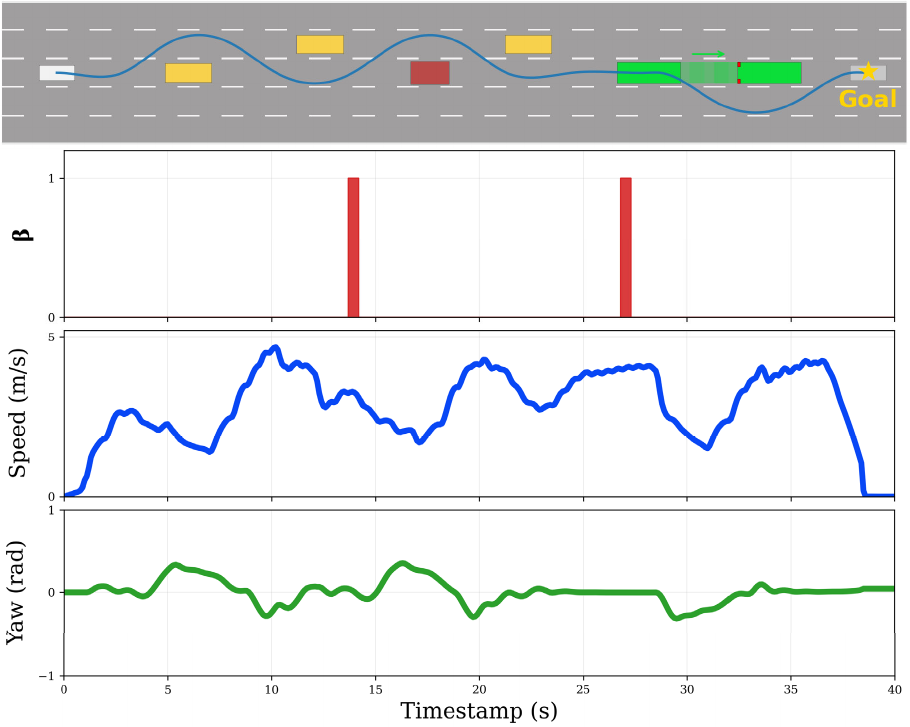}
        \caption{EAMP (Ours)}
    \end{subfigure}

    \caption{Qualitative comparison of trajectory executions facing static obstacles and behavioral anomaly.}
    \label{fig:exp_c_traj}
    \vspace{-10pt}
\end{figure}

\textbf{Qualitative Comparison.}
Fig.~\ref{fig:exp_c_traj} visualizes a representative execution. As indicated by the trigger signals (red bars), both methods successfully activate semantic reasoning upon encountering the unknown static obstacle at $t \approx 14$\,s, confirming OCP effectively resolves its own baseline configuration. However, a critical divergence occurs during the newly introduced dynamic anomaly. When the recognized leading vehicle abruptly brakes, OCP's object-level perception confidence remains rigidly high. Therefore, it structurally fails to generate a secondary response, forcing reliance on myopic geometric braking that causes its speed to plummet to zero (stagnation). In contrast, PC-SET directly evaluates interaction dynamics rather than mere visual familiarity. It successfully identifies the hazardous deceleration, generating a crucial secondary trigger at $t \approx 27$\,s. This timely intervention instantly reconfigures the controller for a proactive lane change, preserving momentum and averting collision.

%--------------------------------------------------
\begin{table}[t]
\vspace{3.5pt}
\centering
\caption{Quantitative comparison under static obstacle and behavioral anomaly scenarios (20 trials).}
\label{tab:exp_c_metrics}
\resizebox{\columnwidth}{!}{
\begin{tabular}{lcccc}
\toprule
\textbf{Method} &
\textbf{SRate} (\%) $\uparrow$ &
\textbf{FTime} (s) $\downarrow$ &
\textbf{TLen} (m) $\downarrow$ &
\textbf{SVar} (m/s) $\downarrow$ \\
\midrule
OCP & 60 & 38.07 & 114.18 & 1.34 \\

\rowcolor{blue!8}
EAMP & \textbf{95} & \textbf{37.85} & \textbf{113.29} & \textbf{1.32} \\
\bottomrule
\end{tabular}}
\end{table}
%--------------------------------------------------

%--------------------------------------------------
%--------------------------------------------------

\section{Conclusion}
In this work, we propose EAMP, an event-adaptive semantic navigation framework for safety-critical dynamic interactions. Unlike methods relying on rigid intervals or object-level tracking, PC-SET explicitly detects behavioral anomalies to asynchronously trigger reasoning. Driven by the distilled SemNav-VLM, the SMPC module systematically maps these strategy-level decisions to structured objective reconfigurations, safely decoupling \emph{when} to reason from \emph{how} to adapt. Extensive experiments in CARLA demonstrate that EAMP significantly outperforms existing baselines by securing superior safety and efficiency without excessive query latency. Future work will focus on deploying this framework on physical mobile robots and extending event-adaptive reasoning to richer real-world interaction scenarios.

\bibliographystyle{IEEEtran}
\bibliography{root}

\end{document}